\pdfoutput=1

\documentclass[11pt]{article}

\usepackage[final]{acl}

\usepackage{hyperref}
\usepackage{url}

\usepackage{pifont}
\usepackage{caption, subcaption}
\usepackage{colortbl}
\usepackage{adjustbox}
\usepackage{float}
\usepackage{makecell}
\usepackage{booktabs}
\usepackage{color}
\usepackage{arydshln}
\usepackage{multirow}
\usepackage{url}
\usepackage{mathtools}
\usepackage{amssymb,enumitem}
\usepackage{times}
\usepackage{latexsym}
\usepackage{times}
\usepackage{latexsym}
\usepackage{tcolorbox}

\usepackage[T1]{fontenc}

\usepackage[utf8]{inputenc}

\usepackage{microtype}

\usepackage{inconsolata}

\title{Divide-or-Conquer? Which Part Should You Distill Your LLM?}

\author{Zhuofeng Wu$^\dagger$$^\ddag$\thanks{Work done during internship at Apple}, He Bai$^\ddag$, Aonan Zhang$^\ddag$,\\
{\bf Jiatao Gu$^\ddag$, VG Vinod Vydiswaran$^\dagger$, Navdeep Jaitly$^\ddag$, Yizhe Zhang$^\ddag$} \\
$^\dagger$University of Michigan, $^\ddag$Apple \\
{\tt \{zhuofeng,vgvinodv\}@umich.edu}, \\ 
{\tt \{hbai22,aonan\_zhang,jgu32,njaitly,yizhe\_zhang\}@apple.com}
}

\begin{document}
\maketitle
\begin{abstract}
Recent methods have demonstrated that Large Language Models (LLMs) can solve reasoning tasks better when they are encouraged to solve subtasks of the main task first.
In this paper we devise a similar strategy that breaks down reasoning tasks into a \textit{problem decomposition} phase and a \textit{problem solving} phase and show that the strategy is able to outperform a single stage solution. 
Further, we hypothesize that the decomposition should be easier to distill into a smaller model compared to the problem solving because the latter requires large amounts of domain knowledge while the former only requires learning general problem solving strategies.
We propose methods to distill these two capabilities and evaluate their impact on reasoning outcomes and inference cost.
We find that we can distill the problem decomposition phase and at the same time achieve good generalization across tasks, datasets, and models.
However, it is harder to distill the problem solving capability without losing performance and the resulting distilled model struggles with generalization.
These results indicate that by using smaller, distilled problem decomposition models in combination with problem solving LLMs we can achieve reasoning with cost-efficient inference and local adaptation.
\footnote{Our code is publicly available at \url{https://github.com/apple/ml-divide-or-conquer}.}

\end{abstract}

\section{Introduction}
\begin{figure*}[!ht]
	\centering
	\begin{center}
		\includegraphics[width=0.81\textwidth]{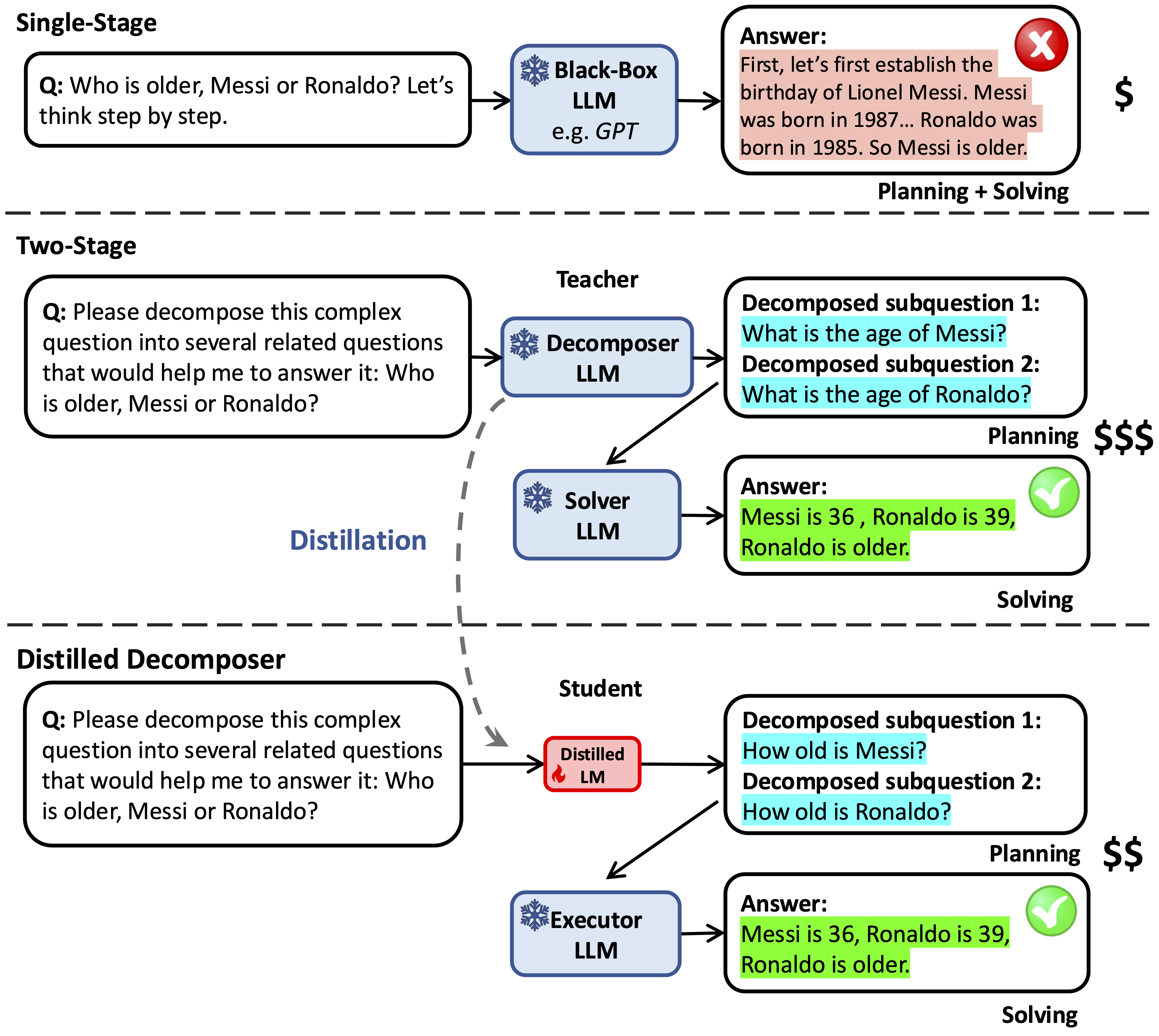}
		\caption[7.5pt]{
Reasoning with a long thought chain using the black box LLM can be expensive and inflexible. We propose to dissect the decomposition and solving of the task, and distill only the decomposition capability to a less costly and more flexible student model, while still maintaining the original performance.
        }
		\label{fig:overview}
		\vspace{-0.2in}
	\end{center}
\end{figure*}

Large Language Models (LLMs), such as GPT-4 \citep{OpenAI2023GPT4TR}, demonstrate exceptional abilities in solving knowledge-intensive tasks like Open Domain QA (ODQA) \citep{zhu2021retrieving}, math \citep{yue2023mammoth}, science \citep{taylor2022galactica} and autonomous agents \citep{yao2022react, Auto-GPT, wang2024executable}. 
However, the use of gigantic LLMs with hundreds of billions of parameters can be \textit{costly} during inference, particularly when the reasoning chain generated is lengthy.
Additionally, due to the opaque nature of these black box LLMs, they offer limited \textit{adaption} options.
There is a need to use \textit{cheaper} and more \textit{flexible} models to leverage the power of these black box LLMs for local adaptation and \textit{cost-efficient} inference.
Distilling the large LLMs would seem like a reasonable strategy, but it often results in a significant drop in performance for downstream tasks \citep{vicuna2023}.

Previous studies \citep{weng2023agent,wang2023survey} have indicated that effectively addressing such tasks requires the model to proficiently perform two essential capabilities simultaneously: 1) \textbf{planning and decomposition}, which involves breaking down complex objectives into smaller, more manageable subgoals to facilitate efficient handling of intricate tasks; and 2) \textbf{execution and solving}, which involves memorizing vast amounts of knowledge from extensive web training data and effectively recalling this information when needed to execute the problem-solving process. The first capability, decomposition, typically requires the model to engage in self-reflection on the input query and generate a Chain-of-Thoughts (CoT)-style reasoning chain \citep{wei2022chain} to tackle the problem. 
Usually, these two abilities are intertwined in a single monolithic model throughout the problem-solving process \citep{zhou2022least}. 

In this paper, we first investigate whether it is possible to decouple the decomposition and solving capabilities, and how to distill these capabilities into smaller models for faster inference. We then test several hypotheses: 1) \textit{Is distilling decomposition easier than distilling solving?} Decomposition primarily relies on semantic understanding and query parsing, while solving requires more domain expertise and knowledge. For example, decomposing the query ``\textit{who is older, Messi or Ronaldo?}'' into ``\textit{how old is Messi?}'', ``\textit{how old is Ronaldo}?'', and ``\textit{who is older?}'' only requires text comprehension, whereas solving the task necessitates memorization, retrieval, and utilization of information. We speculate that compressing the less \textit{knowledge-intensive} decomposition is easier. 2) \textit{Is decomposition capability more generalizable than solving capability?} We hypothesize that decomposition can sometimes be abstracted into symbolic principles, making it more universally applicable across tasks, datasets, and models. This enables tasks and models to share a common decomposition engine and benefit from each other's power, reducing the effort and costs involved in distilling a model for each individual task.

A natural question arises: \textit{is it possible to distill only the long reasoning chain, which accounts for most of the inference cost, but is relatively easier to distill?} To this end, we propose and evaluate the distillation of only the decomposition capability from the LLM. We empirically verified our hypothesis using a teacher model, GPT-3.5-turbo, and two student models, Vicuna-13B~\citep{vicuna2023} and Mistral-7B~\citep{jiang2023mistral}, on QA, mathematics, and compositional datasets~\citep{dua-etal-2019-drop,cobbe2021training,press2022measuring}. \\Our contributions include:
\begin{enumerate}
    \item We demonstrate that the decomposition capability is crucial for complex reasoning of LLMs. This capability can be dissected from the problem solving or task solving capability.
    \item We show the feasibility and effectiveness of distilling only the query decomposition from the teacher model. The resulting distilled model maintains most of the performance while significantly reducing inference costs. In contrast, distilling the solving component of the LLM leads to a considerable decline in performance.
    \item We illustrate that the distilled query decomposition model exhibits good generalization across tasks, datasets, and models. However, the distilled solving ability does not generalize well.
    \item We demonstrate that a static decomposition and solving framework provides more advantages compared to a dynamic approach in which planning and solving steps are interdependent.
\end{enumerate}

\section{Decoupling Decomposition and Solving}
\label{sec:decouple}
As shown in Figure~\ref{fig:overview}, a common approach to solving a reasoning task using an LLM involves directly generating a response to the instruction and question. This is referred to as the \textbf{Single-Stage} model. The conventional method for LLM, known as the Chain of Thought (CoT), instructs the model to ``think step by step,'' allowing the model to take more computational steps for difficult tasks.

However, CoT-style reasoning has limitations as it often struggles to generalize to problems beyond the scope of the in-context examples. To address this drawback, the most notable work is the \textit{Least-to-Most} approach \citep{zhou2022least}, where the model breaks down the original question into subquestions and answers them sequentially. These approaches have shown improved performance compared to CoT.

For QA tasks, typically, the next subquestion is less dependent on the answer to the previous subquestions. Conveniently, we propose a \textit{static} strategy similar to HuggingGPT \citep{shen2023hugginggpt}, where in the first \textbf{Decomposition} stage several decomposed subquestions are first generated to decompose the primary question. In the second \textbf{Solving} stage, these subquestions are then answered one by one to obtain the final answer. We refer to this line of models as the \textbf{Two-Stage} models. 

\textbf{Interactive vs static process} Note that an \textit{interactive} and dynamic process could be beneficial for certain reasoning tasks.
In our experiments with math and QA datasets, the decomposition and solving stages are more independent, thus we did not observe gain by switching to an interactive process. Our primary focus lies in understanding the impact of distilling task decomposition and solving capabilities, rather than finding the optimal framework. Using a \textit{static} approach would enable us to have a clearer separation of the decomposition and solving. The distilled decomposer can also potentially be integrated into more dynamic reasoning processes, enabling iterative solving and refinement based on intermediate outputs. We include a comparison in Section~\ref{abl-decom} to illustrate the benefits of adopting a static process.

\section{Distill the Decomposition Capability}
Generating decomposed questions can be computationally expensive when the reasoning chain is long while using a black box LLM. Moreover, it is challenging to optimize or customize the decomposition process as it is performed by the black box model.
Our proposal aims to address these issues by utilizing a smaller trainable student model, as a drop-in replacement for the large black box LLM for decomposition. To achieve this, we distill the decomposition capability from the teacher LLM, referred to as $\mathcal{T}$.

\paragraph{Generating Sub-questions from Teacher} As shown in Figure~\ref{fig:overview}, we begin by gathering demonstrations from $\mathcal{T}$. Instead of requesting $\mathcal{T}$ to solve the problem, we ask it to break down a given question $Q$ without providing the solution. Specifically, we provide $\mathcal{T}$ with an instruction for decomposition, denoted as $I_\text{decomp}$, along with $Q$. $\mathcal{T}$ then generates a set of sub-questions $\{S_i\}_{i=1,2,3...}$.

\begin{center}
\begin{tcolorbox}[title=Instruction for decomposition: $I_\text{decomp}$, width=0.49\textwidth]
{
\small
Your task is to break down a given complex question into the most relevant and helpful subquestions, ensuring that no more than three subquestions are formulated for each question. Both the context and the main question will be provided to you. If the question does not need breaking down to be answered, return ``No decomposition''; otherwise, list the necessary subquestions. Only return subquestions that directly aid in answering the original question, avoiding any that could be harmful or unhelpful.

Question: \textcolor{blue}{$Q$}
}
\end{tcolorbox}
\end{center}

\paragraph{Decomposer Distillation} 
\label{sec:stage1}
Given the sub-questions $\{S_i\}$ generated from the teacher, we can finetune a student decomposer $\mathcal{S}$ by optimizing the cross-entropy loss for $\mathcal{T}(I_\text{decomp},Q) \rightarrow \{S_i\}$. 
We denote the resulting student model as $\mathcal{S}_D$-$\mathcal{T}$.

\paragraph{Subquestions Screening via Ground-truth Answer}
As an additional step, if the dataset comes with a corresponding ground-truth answer, denoted as $A$, we can optionally use this information to screen high-quality generated subquestions. To do this, we feed the same teacher model $\mathcal{T}$ with another instruction $I_\text{ans}$ that asks the model to solve the primary question $Q$ by first solving the subquestions $\{S_i\}$. We collect the generated answer $\mathcal{T}(I_\text{ans}, P, \{S_i\},Q) \rightarrow \hat{A}$, where $P$ represents the premise. 
$I_\text{ans}$ is provided as the following:
\begin{center}
\begin{tcolorbox}[title=Instruction for solving: $I_\text{ans}$, width=0.49\textwidth]
{
\small
Solve a complex question by answering several related subquestions that would help me to answer it first. Answer the subquestions one by one and finally solve the original question. The final answer is supposed to attached in the end in the format of ``The answer is: ''. Now comes our primary question and its subquestions: \\

Premise: \textcolor{blue}{$P$} \\
Question: \textcolor{blue}{$Q$} \\
SubQuestion: \textcolor{blue}{$\{S_i\}$}

}
\end{tcolorbox}
\end{center}

We assume that, statistically speaking, better $\{S_i\}$ will eventually lead to better resolving the tasks. Thus, we can optionally filter out training instances where $\hat{A} \neq A$. However, this will result in data loss. As this screening process is similar to the \textit{Rejection Sampling} \citep{touvron2023llama2}, we denote the resulting model as $\mathcal{S}_D$-$R$.

In Section~\ref{sec:main}, we compare the performance of the distilled decomposer trained using the entire set of demonstrations $\mathcal{S}_D$-$\mathcal{T}$ against decomposer trained using a screened dataset $\mathcal{S}_D$-$R$.

\begin{table*}[tb]
\centering
\small
\begin{tabular}{p{7em}cccccccc}
\toprule
& \textbf{Decomposer} & \textbf{Solver} & \multicolumn{3}{c}{\textbf{Performance}$\uparrow$} & \multicolumn{3}{c}{\textbf{Inference Expense} $\downarrow$} \\
\cmidrule(lr){2-2} \cmidrule(lr){3-3} \cmidrule(lr){4-6} \cmidrule(lr){7-9}
& Model & Model & GSM8K & DROP & Bamb & GSM8K(\$) & DROP(\$) & Bamb(\$) \\
\midrule
\multirow{2}{*}{\textit{Single-stage}}
& - & GPT &  20.32 & 46.51 & 49.6 & -/0.01 & -/0.05 & -/7e-3\\
& - & Vicuna-13B & 9.40 & 26.68 & 33.6 &-/0.03& -/0.03 &\textbf{-/2e-3}\\
\midrule
\multirow{4}{*}{\textit{Two-stage}}
& GPT & GPT & 65.13 & 55.73 & \textbf{54.4} & 0.13/0.63 & 0.73/0.96 & 2e-3/9e-3\\
& Vicuna-13B & GPT & 62.93 & 47.13 & 48.8 & 0.02/0.67 & 0.07/0.96 & 1e-3/0.02\\
& GPT & Vicuna-13B & 28.13 & 21.29 & 32.8 & 0.13/0.07 & 0.73/0.08 & 2e-3/3e-3\\
& Vicuna-13B & Vicuna-13B & 28.51 & 20.90 & 29.6 & 0.02/0.08
& 0.07/0.08 & 1e-3/5e-3\\
\midrule
\multirow{2}{*}{\textit{w/o oracle answer}}
& $\mathcal{S}_D$-$\mathcal{T}$ & GPT & 67.02 & 55.19 & 52.0 & 0.01/0.62&\textbf{0.06/0.96} & 1e-3/0.03\\
& GPT & $\mathcal{S}_E$-$\mathcal{T}$ & 48.98 & 13.37 & 31.2 & 0.13/0.09 & 0.73/0.06 & 2e-3/2e-3\\
\midrule
\multirow{2}{*}{\textit{w/ oracle answer}}
& $\mathcal{S}_D$-$R$ & GPT & \textbf{67.78} & \textbf{57.97} & 52.0 & \textbf{0.01/0.60} & 0.06/1.11 & 8e-5/7e-3\\
& GPT & $\mathcal{S}_E$-$A$ & 51.55 & 20.34 & 40.8 & 0.13/0.09 & 0.73/0.04 & 2e-3/2e-3\\
\bottomrule
\end{tabular}
\caption{
    Comparison results on GSM8K, DROP, and Bamboogle datasets. Performance on GSM8K is assessed via the exact match score (EM), DROP is evaluated using the F1 score, Bamboogle (Bamb) is evaluated using accuracy. The inference expense is estimated by total sample cost. $X/X$ indicates decomposition/solving cost. 
}
\label{tab:main}
\end{table*}

\section{Experiments}
\textbf{Datasets} We assess the effectiveness of our pipeline on three distinct datasets. GSM8K~\citep{cobbe2021training} focuses on mathematical reasoning and is composed of 7.5K training instances alongside 1K test problems. DROP~\citep{dua-etal-2019-drop} caters to Question Answering, containing 77.4K training samples and a 9.5K validation set. Bamboogle~\citep{press2022measuring} is a manually crafted collection that integrates two questions into one complex question, consisting of 125 test samples. We use the GSM8K test set, the DROP development set, and the Bamboogle test set for evaluation.

\paragraph{Teacher/Student Models} We use GPT-3.5-Turbo-0615 model \citep{ouyang2022training} as the teacher model throughout our experiments.  
After training we employ different levels of teacher models to ensure a comprehensive evaluation: two open sourced models (Vicuna-13b-v1.3 and Mistral-7B-Instruct-v0.3) and three black box models (text-davinci-003~\citep{brown2020language}, GPT-3.5-Turbo and GPT-4). 

\paragraph{Student solver Models}
To compare the performance of distilling decomposer with distilling solver, we conducted further training on several Vicuna models to mimic the behavior of the teacher as student solvers. Similar to the student decomposer, $\mathcal{S}_E$-$\mathcal{T}$ represents the model trained using the teacher's demonstrations of $\mathcal{T}(I_\text{ans},\{S_i\},Q) \rightarrow (\{\hat{A^s_i}\},  \hat{A})$, where $\{\hat{A^s_i}\}$ represents the answers to the subquestions $\{S_i\}$ generated by $\mathcal{T}$.

Furthermore, in scenarios where the oracle answer $A$ is available, we fine-tuned the same vanilla Vicuna-13B model to obtain $\mathcal{S}_E$-$A$. This model was trained using $(I_\text{ans},\{S_i\},Q) \rightarrow (\{\hat{A^s_i}\}, A)$, where the targets include answers to the subquestions $\{S_i\}$ from the $\mathcal{T}$ and the oracle answer $A$.

\paragraph{Training Details} 
We use a batch size of 128, train for 3 epochs on DROP and train for 5 epochs on GSM8K and Bamboogle dataset (until convergence), and set the learning rate to $2\cdot10^{-5}$ for the distillation training. All the distillation fine-tuning can be finished in less than 12 hours on 8 $\times$ 80G A100 GPUs.

\paragraph{Inference Cost Estimation} 
We calculate the cost based on GPT-3.5-turbo-1106 (175B - a vague estimation based on GPT-3), with a rate of $\$0.001$ for 1000 input tokens and $\$0.002$ for 1000 output tokens.
OpenAI has made significant optimizations for inference time when serving GPT models. To ensure a fair comparison, we conservatively estimate the cost of the Vicuna-13B model by dividing the cost by the ratio of the model size. As a result, the cost for Vicuna-13B is approximately $\$7.42*10^{-5}$ for 1000 input tokens and $\$1.48*10^{-4}$ for 1000 output tokens.

\section{Results}
\subsection{Decomposition is Essential for Reasoning}
First, we explore the possibility of separating the \textit{Decomposition} from \textit{Solving} and assess the effectiveness of using an improved decomposition for complex reasoning tasks.

Previous studies~\citep{press2022measuring, zhou2022least} have demonstrated the utility of leveraging decomposed subquestions to enhance the question-answering capabilities of black-box models. They
adopt \textit{interactive} planning strategies, where the generation of each subquestion is conditioned on the answer of the previous subquestions.

As discussed in Section~\ref{sec:decouple}, we instead use a \textit{static} strategy by breaking down the reasoning process into two separate stages of Decomposition and Solving. Table~\ref{tab:main} (Single-stage GPT/Vicuna vs Two-stage GPT/Vicuna), shows that in general such a static strategy leads to performance gains over a Single-stage approach. This aligns with previous findings \citep{zhou2022least}.

We demonstrate in Table~\ref{tab:main} (Two-stage models) that replacing a stronger decomposer (GPT) with a weaker decomposer (Vicuna) mostly results in a noticeable decrease in performance, with an exception of using Vicuna as solver on GSM8K. We hypothesize that presumably the Vicuna solver is too erroneous to harness the improvement from the decomposition. We observe that the decrease is more significant when the solver is more powerful.
This suggests that in order to achieve optimal performance, a stronger decomposer is essential.

\subsection{Is Distilling Decomposition Easier than Distilling Solving?} \label{sec:main}
Next, we investigate distilling knowledge from $\mathcal{T}$ to $\mathcal{S}$ when the ground truth answer $A$ is not available. This is the most common use case as ground truth annotations are typically expensive and rare. The results are shown in Table~\ref{tab:main} (w/o oracle answer $A$). It can be seen that swapping in $\mathcal{S}_D$-$\mathcal{T}$ for the decomposer is at least comparable to the performance using $\mathcal{T}$. Moreover, the $\mathcal{S}_D$-$\mathcal{T}$ exhibits a noticeable improvement compared to using Vicuna as the decomposer.
However, swapping in a student solver model $\mathcal{S}_E$-$\mathcal{T}$ significantly harms the performance. 
We also evaluated a single-stage student model distilled from single-stage GPT. The result, omitted, was even worse than the model where GPT was the decomposer and $\mathcal{S}_E$-$\mathcal{T}$ was the solver.
In terms of inference cost, our $\mathcal{S}_D$-$\mathcal{T}$ approach results in significantly lower cost for the decomposition compared to using the teacher GPT model. The cost of the solver remains unchanged.

We compare some decompositions from $\mathcal{T}$, from Vicuna and from $\mathcal{S}_D$-$\mathcal{T}$ on the evaluation set in Table~\ref{tab:qual}. It can be observed that the distilled $\mathcal{S}_D$-$\mathcal{T}$ model, which is obtained by using in-domain demonstration from $\mathcal{T}$, exhibits a high degree of similarity to the teacher demonstration in the generated subquestions on the unseen test set. In contrast, the original Vicuna model often generates unhelpful questions that have the potential to distract the solver (Table~\ref{tab:qual}, Vicuna-13B).

One might naturally wonder: \textit{If a smaller student model can quickly imitate the decomposition abilities of the teacher model, why is it challenging to acquire this skill directly through student model's initial pretraining?}
Our hypothesis is that the decomposition ability of a stronger teacher model is easy to distill but difficult to acquire. 
This skill is likely based on the thorough digestion and internalization of vast amounts of data during the intensive pretraining of the larger models. 
However, as it is more logical and abstract rather than being \textit{knowledge-intensive}, a few demonstrations may already provide ample guidance to the student. To draw an imperfect analogy, finding a physics theorem from massive observations is much more challenging than learning the theorem in the class.

\begin{table*}[ht!]
\small
\centering
\renewcommand{\arraystretch}{1.5}

\begin{tabular}{|p{4.4cm}|>{\centering\arraybackslash}p{1.85cm}|p{8.4cm}|}
    \hline
    \rowcolor{gray!25}
    \textbf{Dataset: DROP} & \textbf{Models} & \textbf{Decomposed Sub-questions} \\
    \hline
    \multirow{4}{4.4cm}{\textbf{Premise $P$}:  The Raiders stayed at home for a Week 16 duel with the Houston Texans. ... The Texans tried to rally in the fourth quarter as Brown nailed a 40-yard field goal, yet the Raiders' defense would shut down any possible attempt.\\\textbf{Question $Q$}: How many field goals did both teams kick in the first half?} & Vicuna-13B & \makecell[l]{1. Which teams played against each other? \textcolor{red}{X}\\
2. What were the scores for each team during the game? \textcolor{red}{X}\\
3. Which team had the lead at the end of the game? \textcolor{red}{X}
} \\
    \cline{2-3}
     & GPT-3.5 & \makecell[l]{1. How many field goals did the Raiders kick in the first half?\\
2. How many field goals did the Texans kick in the first half?\\
3. What is the sum of the field goals kicked by both teams in \\ the first half?\\
} \\
    \cline{2-3}
     & \makecell[c]{$\mathcal{S}_D$-$\mathcal{T}$\\\textit{In-Domain}} & \makecell[l]{1. How many field goals did the Raiders kick in the first half? \\
2. How many field goals did the Texans kick in the first half? \\
} \\
    \cline{2-3}
     & \makecell[c]{$\mathcal{S}_D$-$\mathcal{T}$(GSM)\\\textit{Cross-Domain}} & \makecell[l]{1. How many field goals did the Raiders kick in the first half? \\
2. How many field goals did the Texans kick in the first half?
} \\
    \hline
\end{tabular}

\begin{tabular}{|p{4.4cm}|>{\centering\arraybackslash}p{1.85cm}|p{8.4cm}|}
    \hline
    \rowcolor{gray!25}
    \textbf{Dataset: GSM8K} & \textbf{Models} & \textbf{Decomposed Sub-questions} \\
    \hline
    \multirow{4}{4.4cm}{\textbf{Premise $P$}: Mark is a copy-editor. He edits an equal number of sentences each week for two different publishers, who each pay him a different rate per sentence. Publisher B pays Mark twice what Publisher A pays. Mark edits a total number of 1000 sentences each week, and Publisher A pays him 5 cents per sentence. \\\textbf{Question $Q$}: How much does Mark make in a week, in cents?} & Vicuna-13B & \makecell[l]{1. What is the rate per sentence that Publisher B pays Mark? \textcolor{red}{X} \\
2. What is the total amount Publisher A pays Mark for editing \\ 1000 sentences? \\
3. What is the total amount Publisher B pays Mark for editing \\ 1000 sentences?
} \\
    \cline{2-3}
     & GPT-3.5 & \makecell[l]{1. How many sentences does Mark edit each week for Publisher A? \\2. How many sentences does Mark edit each week for Publisher B? \\3. How much does Mark make per sentence from Publisher B?} \\
    \cline{2-3}
     & \makecell[c]{$\mathcal{S}_D$-$\mathcal{T}$\\\textit{In-Domain}} & \makecell[l]{1. How many sentences does Mark edit for Publisher A in a week? \\2. How many sentences does Mark edit for Publisher B in a week? \\3. What is the rate per sentence paid by Publisher B?} \\
    \cline{2-3}
     & \makecell[c]{$\mathcal{S}_D$-$\mathcal{T}$(DROP)\\\textit{Cross-Domain}} & \makecell[l]{1: How much does Publisher A pay Mark per sentence? \\
2: How much does Publisher B pay Mark per sentence? \\
3: How many sentences does Mark edit in a week?} \\
    \hline
\end{tabular}

\begin{tabular}{|p{4.4cm}|>{\centering\arraybackslash}p{1.85cm}|p{8.4cm}|}
    \hline
    \rowcolor{gray!25}
    \textbf{Dataset:  Bamboogle} & \textbf{Models} & \textbf{Decomposed Sub-questions} \\
    \hline
    \multirow{4}{4.4cm}{\textbf{Question $Q$}: When was the longest bridge in the world opened?} & Vicuna-13B & \makecell[l]{1. What is the definition of a ""bridge"" for the purpose of \\determining the longest one in the world? \textcolor{red}{X}\\
2. What is the current longest bridge in the world and when \\was it opened?\\
} \\
    \cline{2-3}
     & GPT-3.5 & \makecell[l]{1. What is the name of the longest bridge in the world?\\
2. When was the longest bridge in the world opened?\\
} \\
    \cline{2-3}
     & \makecell[c]{$\mathcal{S}_D\text{-}\mathcal{T}$\\\textit{In-Domain}} & \makecell[l]{1. What is the name of the bridge that holds the record for being \\the longest in the world?  \\
2. What is the date on which this record-breaking bridge was \\officially opened to traffic?  \\
} \\
    \hline
\end{tabular}
\caption{Examples for decomposed subquestions from each method on GSM8K, DROP, and Bamboogle. $\mathcal{S}_D$-$\mathcal{T}$(GSM) and $\mathcal{S}_D$-$\mathcal{T}$(DROP) denote Vicuna student models that distilled from $\mathcal{T}$'s demonstration on GSM8K and DROP datasets, respectively. \textcolor{red}{X} indicates not helpful subquestions. }
\label{tab:qual}
\end{table*}

\paragraph{With available oracle answers}
Sometimes, we have access to the oracle answers $A$, which can be used to further enhance the model's performance on specific domains through local adaptation and additional finetuning. %
As a result, the performance on these target domain can be beyond the performance of the black-box teacher model. We explore the options to enhance the models via distillation or target domain finetuning. 

In these scenarios, we can possibly use $A$ to screen the training instance for distill the decomposer, similar to Rejection Sampling. 
The resulting student model $\mathcal{S}_D$-$R$ achieved higher performance than using $\mathcal{S}_D$-$\mathcal{T}$, as shown in Table~\ref{tab:main} (w/ oracle answer $A$). Notably, on the DROP dataset, $\mathcal{S}_D$-$R$ outperforms the teacher model in F1 score. 

We also finetune another Vicuna model for the solver using the ground-truth answers, referred to as $\mathcal{S}_E$-$A$.
Our main findings remain consistent to the scenario where no oracle answers are available. Distilling the decomposer still yields better performance comparing with finetuning the solver. 
We omitted the single-stage Vicuna model finetuned using $A$, which yielded worse results than GPT(decomposer) + $\mathcal{S}_E$-$A$(solver).

\paragraph{Failure modes for $\mathcal{S}_E$ models} 
According to our observations, we hypothesize that there are two primary failure modes of the $\mathcal{S}_E$-$\mathcal{T}$ and $\mathcal{S}_E$-$A$ models. 

First, answering either subquestions or primary questions would require extensive world knowledge, which can be difficult to compress into a student model that is hundreds of times smaller, using only a few demonstrations. In other words, a strong solving capability is knowledge-intensive. On the other hand, decomposition capability might be more compressible as it is typically more abstract, has lower information density, and is more universal than solving capability. 

Second, since we used the teacher's answers to the subquestions $\{\hat{A^s_i}\}$ as part of the target, the $\mathcal{S}_E$ models could get confused and generate the final answers to one of the subquestions $\{S_i\}$, rather than the primary question $Q$. (Examples are provided in Appendix~\ref{sec:example1}.)

Based on above findings, we excluded the $\{\hat{A^s_i}\}$ in the target when training the $\mathcal{S}_E$ models. Specifically, we train the models to directly generate the answer by skipping answering subquestions, $\mathcal{S}_E$$(I'_\text{ans},\{S_i\},Q) \rightarrow \hat{A}~\text{or}~A$. The resulting models are denoted as $\mathcal{S}_E$-$\mathcal{T}$(direct) and $\mathcal{S}_E$-$A$(direct) in Table~\ref{tab:as}. We found that $\{\hat{A^s_i}\}$ from the target yields improved results over the DROP dataset, but leads to a decrease in performance over the GSM8K dataset. Overall, the decrease observed in GSM8K is more prominent than the gain seen in the DROP dataset. Therefore, we still use the $\mathcal{S}_E$ models with the $\{\hat{A^s_i}\}$ in the target. We provide additional analysis, $I'_\text{ans}$, and show the comparison results in Appendix~\ref{app:as}.

\begin{table}[ht]
\centering
\small
\begin{tabular}{cp{1cm}p{1cm}p{1cm}p{1cm}}
\toprule

\textbf{Decomposer} & GPT & $\mathcal{S}_D$-$R$ & {GPT} & {\textbf{-}}   \\
\textbf{Solver} & GPT & {GPT} & $\mathcal{S}_E$-$A$ & $\mathcal{S}_E$-$A$   \\

 \cmidrule(lr){1-5} 
   Trained on & \multicolumn{4}{c}{\textbf{Evaluation on DROP}}   \\
 \cmidrule(lr){2-5} 

 GSM8K & 55.73 & 51.05 & 7.98 & 17.22 \\ 
 \midrule
  \midrule
  Trained on & \multicolumn{4}{c}{\textbf{Evaluation on GSM8K}}   \\
 \cmidrule(lr){2-5} 
DROP & 65.13 & 63.15 & 11.30 & 3.41 \\
\bottomrule
\end{tabular}
\caption{
    Distilled student decomposers demonstrate strong generalization over out-domain datasets. 
}
\label{tab:cross}
\end{table}

\begin{table}[ht]
\footnotesize
\centering

\begin{tabular}{p{2cm}ccc}
    \toprule
    \textbf{Decomposor} & \textbf{Solver} & GSM8K & DROP  \\\hline

\multirow{3}{*}{GPT-3.5-Turbo} 
& Vicuna-13B & 28.0 & 33.78 \\ 
& GPT-3.5-Turbo & 66.0 & 59.38 \\
& GPT-4 & 90.5 & 77.60 \\
\midrule

\multirow{3}{*}{Vicuna-13B} 
& Vicuna-13B & 29.5 & 26.56 \\ 
& GPT-3.5-Turbo & 57.0 & 47.31 \\
& GPT-4 & 88.5 & 79.40 \\
\midrule
\multirow{3}{*}{$\mathcal{S}_D$-$R$} 
& Vicuna-13B & 31.5 &  33.38 \\ 
& GPT-3.5-Turbo & 66.5 & 61.94 \\
& GPT-4 & 91.5 & 81.02 \\
    \bottomrule
\end{tabular}

\caption{
    Distilled student decomposers demonstrate consistent improvements over different solvers. Weaker solvers receive more gain. 
}
\label{tab:cross-model}
\end{table}

\subsection{Is Distilling Decomposition More Generalizable than Distilling Solving?}

\paragraph{Generalization to other domains}
We then investigate whether the distilled decomposer, which is trained on a specific domain dataset, can be applied to out-of-domain datasets with distinct objectives. To test this, we perform a cross-domain evaluation on DROP and GSM8K, which require different expertise from the solver. The results, when the oracle answer is available, are presented in Table~\ref{tab:cross}. Surprisingly, the distilled decomposer $\mathcal{S}_D$-$R$ demonstrates good generalization and versatility to the other domain, as evidenced by only a slight decrease in performance compared to using the teacher GPT model as the decomposer. In contrast, when substituting the solver with $\mathcal{S}_E$-$A$, which is fine-tuned on the original domain, the generalization to the other domain is poor regardless of the decomposer used. Some examples of cross-domain subquestion decomposition are shown in Table~\ref{tab:qual}. 
The results on the scenario with no oracle answer are consistent with Table~\ref{tab:cross}.

\paragraph{Generalization to other solvers}
Next, we examine whether the distilled decomposer is compatible and universally suitable for different solvers. The results can be seen in Table~\ref{tab:cross-model}. The performance of $\mathcal{S}_D$-$R$ 
is comparable to that of the teacher decomposer (GPT), and it shows overall improvements over a weaker decomposer (Vicuna) when connected to different solvers. We found that weaker solvers receive more performance gain compared to strong solvers, through upgrading to a distilled decomposer. We hypothesize that the reason lies in the fact that the weaker solver may be incapable of fully utilizing the benefits of the decomposition.

\paragraph{Generalization to other backbones} To verify whether our observations on the Vicuna-13B model hold for other backbone models, we conducted similar experiments on the Bamboogle dataset using the Mistral-7B model (Table~\ref{tab:mistral}). The results exhibited a consistent pattern across different backbones, suggesting the robustness of our conclusions.

\begin{table}[tb]
\centering
\small
\begin{tabular}{ccc}
\toprule
Backbone  & Vicuna-13B & Mistral-7B \\
\midrule
-/Backbone & 33.6 & 40.0 \\
Backbone/GPT & 48.8 & 56.8 \\
GPT/Backbone & 32.8 & 40.0 \\
Backbone/Backbone & 29.6 & 38.4 \\
\midrule
$\mathcal{S}_D$-$\mathcal{T}$/GPT & 52.0  & 55.2 \\
GPT/$\mathcal{S}_E$-$\mathcal{T}$ & 31.2 & 40.8 \\
\midrule
$\mathcal{S}_D$-$R$/GPT & 52.0 & 60.0 \\
GPT/$\mathcal{S}_E$-$A$ & 40.8  & 47.2 \\
\bottomrule
\end{tabular}
\caption{
    Comparison results of Vicuna-13B and Mistral-7B on Bamboogle dataset. $X/X$ denotes the Decomposer/Solver model, where ``Backbone'' refers to the vanilla untuned backbone model.
}
\label{tab:mistral}
\end{table}

\section{Ablations}
We provide an extensive evaluation of various instructions, and an exploration into the influence of the number of demonstrations in Appendix~\ref{sec:abl-ins}.

\subsection{Decomposition: Static or Dynamic?} \label{abl-decom}
Prior research, including self-ask~\citep{press-etal-2023-measuring} and CoQ~\citep{zhu-etal-2023-chain}, employed dynamic decomposition pipelines where each reasoning step depends on previous steps. In contrast, our approach adopts a static method that separates planning from question-solving. This section presents a direct comparison of both pipelines.

\paragraph{Dynamic Pipeline}
Our pipeline is compared to the self-ask dynamic decomposition method. Self-ask appends "Are follow-up questions needed here?" to the original query, determining if subquestions are needed to assist in answering. When needed, the model generates and independently answers relevant subquestions. This process iterates until the model deems no additional subquestions are needed to answer the original query.

\paragraph{Setup}
GPT-3.5-turbo is used for both question decomposition and answer generation in both the dynamic and static model. We compare the two approaches on Bamboogle and GSM8K datasets.

\begin{table}[ht]
\centering
\small
\begin{tabular}{cp{1cm}p{1cm}p{1cm}p{1cm}}
\toprule

 & \multicolumn{2}{c}{\textbf{Bamboogle}} & \multicolumn{2}{c}{\textbf{GSM8K}} \\
 \cmidrule(lr){2-5} 
 & EM & \#tokens & Acc & \#tokens \\

 \cmidrule(lr){1-5} 
   Dynamic &  53.6 & 59,792 & 62.92 & 823,886  \\
 \cmidrule(lr){2-5} 

 Static & \textbf{54.4} & \textbf{15,106} & \textbf{65.13} & \textbf{355,302}\\ 

\bottomrule
\end{tabular}
\caption{
    Comparison between static planning and dynamic planning on Bamboogle and GSM8K datasets. 
}
\label{tab:abl}
\end{table}

\paragraph{Results and Analysis}

From Table~\ref{tab:abl}, we observe that: 
1) Dynamic planning performs poorly on both datasets because dynamic processes require precise reasoning models. Errors in intermediate steps can influence subsequent steps and affect the final outcomes. 2) The cost of the dynamic pipeline is markedly higher than that of the static pipeline ($\times$3.96 and $\times$2.32 for the number of input tokens on Bamboogle and GSM8K, respectively), highlighting the substantial inference costs associated with dynamic processes.

We acknowledge the potential of dynamic decomposition and its advantages in specific application scenarios. However, our experiments on evaluated tasks show that static decomposition offers certain advantages. Nevertheless, our goal is to analyze whether decomposition or solver capability is easier and more generalizable to distill, rather than seeking the optimal strategy.

\subsection{Impact of Learning Decomposition}
Fine-tuning a model to enhance its decomposition performance might affect its problem-solving abilities. To evaluate this effect, we assess the performance of a fine-tuned decomposer against an untrained model in a question-answering context. In the initial question decomposing phase, both models tackle the same set of decomposed questions. Then, during the answering phase, one model leverages an untrained Vicuna model, while the other employs $\mathcal{S}_D$-$\mathcal{R}$, which was initially trained for the decomposition task. We adhered to the evaluation settings in Table~\ref{tab:main}.

\begin{table}[ht]
\centering
\small
\resizebox{0.48\textwidth}{!}{
\begin{tabular}{ccccc}
\toprule
\textbf{Decomposer} & \textbf{Solver} & GSM8K & DROP & Bamb \\
\midrule
$\mathcal{S}_D$-$R$ & Vicuna-13B & \textbf{28.35} & \textbf{26.63} & \textbf{32.8} \\
$\mathcal{S}_D$-$R$ & $\mathcal{S}_D$-$R$ & 5.99 & 7.87 & 17.6 \\ 
\bottomrule
\end{tabular}
}
\caption{
    Impact of learning question decomposition on model's solving capability. 
}
\label{tab:impact}
\end{table}

From Table~\ref{tab:impact}, it is evident that the trained specialized model performs significantly worse than the untuned model on other tasks (-22.36 on GSM8K, -18.76 on DROP, -15.2 on Bamboogle). This indicates that, without specific adjustments or designs, focusing solely on learning decomposition can negatively impact problem-solving abilities in reasoning tasks. These findings underscore our core design principle: separating reasoning into distinct decomposition and solving stages is advantageous. In our pipeline, decomposition process is isolated from the problem-solving stage, ensuring that solver's reasoning capability remains intact.

\section{Related Work}
\textbf{LLM Distillation} 
Tremendous progress ~\citep{jiao-etal-2020-tinybert, sun-etal-2019-patient, li-etal-2021-dynamic} has been made in terms of compressing large-scale pre-trained language models such as BERT~\citep{devlin-etal-2019-bert} or RoBERTa~\citep{liu2019roberta}. For generative models, compression is predominantly achieved by minimizing the K-L divergence between teacher and student distributions~\citep{sanh2019distilbert, gu2023knowledge}. A pivotal assumption underlying these methods is the full accessibility of the teacher model's components. However, most powerful LLMs are black boxes, revealing only limited outputs. Given these constraints, several methodologies have emerged that train directly on data generated by teacher models~\citep{vicuna2023, taori2023stanford}. We follow a similar distillation strategy but focus on the decomposition capability distillation. More recently, researchers have explored aligning small models with large models by leveraging the reward signal from teachers or human preference feedback~\citep{kwon2023reward, gu2023minillm, tunstall2023zephyr}, using methods such as DPO~\citep{rafailov2024direct} or PPO~\citep{schulman2017proximal}. While some approaches, such as Symbolic CoT~\citep{li-etal-2023-symbolic} and SCOTT~\citep{wang-etal-2023-scott}, have focused on distilling reasoning (CoT) capabilities into one compact model, which diverges from our main focus.

\paragraph{LLM Reasoning Chain}
LLMs can benefit from explicit reasoning chains, as demonstrated by recent studies \citep{wei2022chain, zheng2023progressive}. The Chain of Thought (CoT) \citep{wei2022chain} technique has become standard for enhancing model performance on complex tasks. Tree of Thoughts \citep{yao2023tree} decomposes the problem into multiple thought steps and generates multiple thoughts per step, creating a tree structure. The LLM+P approach \citep{liu2023llm+} incorporates an external classical planner for long-horizon planning and translates the plan back into natural language. Theoretical work \citep{feng2023towards} has analyzed why CoT works by using circuit complexity theory. It shows that without CoT, the model size would need to be prohibitively large to achieve the same performance through direct reasoning.

However, CoT-style reasoning is limited by the fact that it often generalizes poorly to problems beyond the scope of the provided in-context examples \citep{zhou2022least}. To address this, some studies have asked LLMs to decompose complex questions into subquestions following the Least-to-Most prompt \citep{zhou2022least}. Others have used the self-ask method to elicit follow-up questions that aid in addressing the original inquiry \citep{press2022measuring}. Our work contributes to this line of research by extending the horizon to cost-efficient inference and generalization across tasks.

\paragraph{Question Decompostion Datasets and Approaches}
A widely recognized dataset for question decomposition in the literature is QDMR~\citep{wolfson-etal-2020-break}. It comprises an ordered list of sub-questions essential for addressing a primary question. Several previous works have been training question decomposers on the QDMR dataset \citep{guo-etal-2022-complex, zhu-etal-2023-chain}. In contrast, some research does not rely on QDMR but employs their uniquely labeled data. For instance, ~\citet{min-etal-2019-multi} recast question decomposition as a span prediction problem and trained their model on a set of 400 labeled questions. Recognizing the challenges associated with obtaining reliable decomposition data, ~\citet{perez-etal-2020-unsupervised} introduced an unsupervised decomposition approach, capitalizing on the similarity between the primary question and 10M potential sub-questions mined for decomposition purposes. Our approach differs from the aforementioned methodologies because we extract the decomposition power solely from the teacher model, without relying on annotated subquestions.

\paragraph{Complement LLMs with Small models}

There have been studies that have emphasized the potential of smaller, task-specific models to complement the predictions of LLM. \citet{xu2023small} explored a framework in which candidates produced by these task-specific models are fed to an LM, with a primary focus on classification tasks. \citet{welleck2022generating} train a smaller model to iteratively improve sequences generated by LMs. \citet{vernikos2023small} have demonstrated that collecting multiple erroneous outputs from LMs and using a small corrector model to unify the generation can significantly reduce errors. Our work can also be seen as developing a smaller decomposer model to activate the best performance of a large-scale LM.

\section{Conclusion}
Our investigation provides a fine-grained examination of the LLM's capability on reasoning tasks, by disentangling the decomposition and solving aspects. Although both capacities are vital for reasoning, we demonstrate that decomposition is less dependent on specific knowledge and thus easier to distill compared to distilling solving capabilities, regardless of the availability of ground truth labels. 
Additionally, the distilled decomposer shows strong generalization abilities across different tasks, datasets and executor/solvers. For future work, it would be interesting to train universal decomposer models using data from various tasks, and explore the use of reinforcement learning to further enhance the decomposer, leveraging the signal from the solver outcome. Another possible direction for future work is to assess the effectiveness of our method in other long-horizon planning tasks, including LLM-powered agent, tool use, and multiturn decision making.

\clearpage

\section*{Limitations}
Our work is built upon several assumptions. First, we assume that the teacher model is capable of breaking down queries effectively. Second, we assume that the student model has the capacity to learn the distilled planning from the teacher model. Lastly, we assume that the tasks involved in our work require long horizon planning capability. If any of these assumptions do not hold true, it would impact the effectiveness of our proposed method.

It is important to note that we have only assessed the effectiveness of our model in the context of math and QA aspects. In order to fully complete our work, it would be necessary to evaluate our model on a broader range of planning tasks. This would include benchmarks related to tool use, LLM agents, and multiturn scenarios. Such evaluations would help verify the versatility and applicability of our proposed method.

\section*{Acknowledgements}
We thank Wang Zhu, Lingxiao Zhao, Simone Conia, Pratyush Maini, Zhenhao Zhang, Jiazhao Li, Fei Sun, members of the University of Michigan's NLP4Health research group, and all anonymous reviewers for helpful discussion and valuable feedback.

\bibliography{coq}
\clearpage
\appendix

\section*{Appendix}
\section{Exclusion of Answers to Subquestions}
\label{app:as}
\begin{table}[h]
\centering
\small

\resizebox{0.48\textwidth}{!}{
\begin{tabular}{p{5.5em}cccc}
\toprule
& \textbf{Decomposer} & \textbf{Solver} & \multicolumn{2}{c}{\textbf{Performance}$\uparrow$} \\
\cmidrule(lr){2-2} \cmidrule(lr){3-3} \cmidrule(lr){4-5}
& Model & Model & GSM8K & DROP \\
\midrule
\multirow{2}{5.5em}{\centering \textit{w/o oracle $A$}}
& GPT & $\mathcal{S}_E$-$\mathcal{T}$(Direct) & 5.46 & 53.17 \\
& GPT & $\mathcal{S}_E$-$\mathcal{T}$ & 48.98 & 13.37 \\
\midrule
\multirow{3}{5.5em}{\centering \textit{w/ oracle $A$}}
& GPT & $\mathcal{S}_E$-$A$(Direct) & 6.44 & 72.55 \\
& GPT & $\mathcal{S}_E$-$A$ & 51.55 & 20.34 \\
& - & $\mathcal{S}_E$-$A$(Direct) & 11.75 & 72.29  \\
\bottomrule
\end{tabular}
}
\caption{
  Excluding answers to subquestions $\{\hat{A^s_i}\}$ from the target yields improved results over the DROP dataset, but leads to a decrease in performance over the GSM8K dataset.
}
\label{tab:as}
\end{table}

We hypothesize that for tasks involving mathematical reasoning, the answers typically necessitate some form of computation, making a step-by-step solution essential. Without this, setting a numerical value as the fine-tuning target almost invariably results in failure. Conversely, DROP, being a reading comprehension dataset, derives a significant portion of its answers directly from the provided text. In such scenarios, including answers to subquestions poses a risk of disrupting the answer distributions.

The instruction for solving, denoted as $I'_\text{ans}$, remain identical to those specified in $I_\text{ans}$. The only difference comes from the fine-tuning target.

\section{Ablation Study over Instruction for Decomposition}
\label{sec:abl-ins}
\begin{table}[ht]
\footnotesize
\centering
\resizebox{0.45\textwidth}{!}{
\begin{tabular}{cccc}
\toprule
\textbf{Decomposer} & \textbf{Solver} & 0-shot & 1-shot \\
\midrule
\multirow{2}{8em}{\centering GPT-3.5-Turbo} 
& GPT-3.5-Turbo & 66.0 & 70.0 \\
& GPT-4 & 90.5 & 91.5 \\
\midrule

\multirow{2}{8em}{\centering Vicuna-13B} 
& GPT-3.5-Turbo & 57.0 & 61.5 \\
& GPT-4 & 88.5 & 91.5 \\
\midrule
\multirow{2}{8em}{\centering $\mathcal{S}_D$-$R$} 
& GPT-3.5-Turbo & 66.5 & 67.5 \\
& GPT-4 & 91.5 & 91.5 \\
    \bottomrule
\end{tabular}
}
\caption{
    Impact of including demonstration in decomposition instruction, examined on a subset of GSM8K dataset. 
}
\label{tab:1-shot}
\end{table}

Prior research has demonstrated that incorporating demonstrations within prompts can significantly enhance the ability of Large Language Models to adhere to given instructions. Our findings in Table~\ref{tab:1-shot} further substantiate this, revealing that including a single-shot demonstration notably improves the quality of decomposed questions. This enhancement has been consistently observed across a variety of decomposers.

\begin{center}
\begin{tcolorbox}[title=Instruction for decomposition: $I_\text{decomp}$, width=0.49\textwidth]
{
Your task is to break down a given complex question into the most relevant and helpful subquestions\textcolor{red}{, ensuring that no more than three subquestions are formulated for each question}. Both the context and the main question will be provided to you. If the question does not need breaking down to be answered, return ``No decomposition''; otherwise, list the necessary subquestions. Only return subquestions that directly aid in answering the original question, avoiding any that could be harmful or unhelpful.

Question: \textcolor{blue}{$Q$}
}
\end{tcolorbox}
\end{center}

We have conducted an ablation study focusing on the instructions used for question decomposition. Our goal is for the resulting subquestions to act as useful cues for the executor, all the while ensuring they do not introduce unnecessary information. Central to our design rationale is determining the optimal number of subquestions the decomposer should produce. 
More specifically, we analyzed outcomes where no restrictions were applied (removing the highlighted part in $I_\text{decomp}$) and compared these against scenarios with varying maximum numbers of subquestions allowed. The results of these investigations are detailed in Table~\ref{tab:abl-qd}. Our findings succinctly reveal that a cap of "no more than three subquestions" yields the most effective results. 
\begin{table}[ht]
\footnotesize
\centering

\begin{tabular}{l:cc}
    \toprule
    \textbf{Instruction} & EM & F1\\
    \midrule
    no restriction & 45.69 & 56.63 \\
    no more than four & 46.40 & 57.19\\
    no more than three & 50.00 & \textbf{59.88}\\
    no more than two & 46.89 & 58.47\\
    \bottomrule
\end{tabular}

\caption{
    Effect of limiting the maximum number of subquestions in decomposition instructions on a subset of the DROP dataset.
}
\label{tab:abl-qd}
\end{table}

\begin{figure*}[!ht]
	\centering
	\begin{center}
		\includegraphics[width=0.98\textwidth]{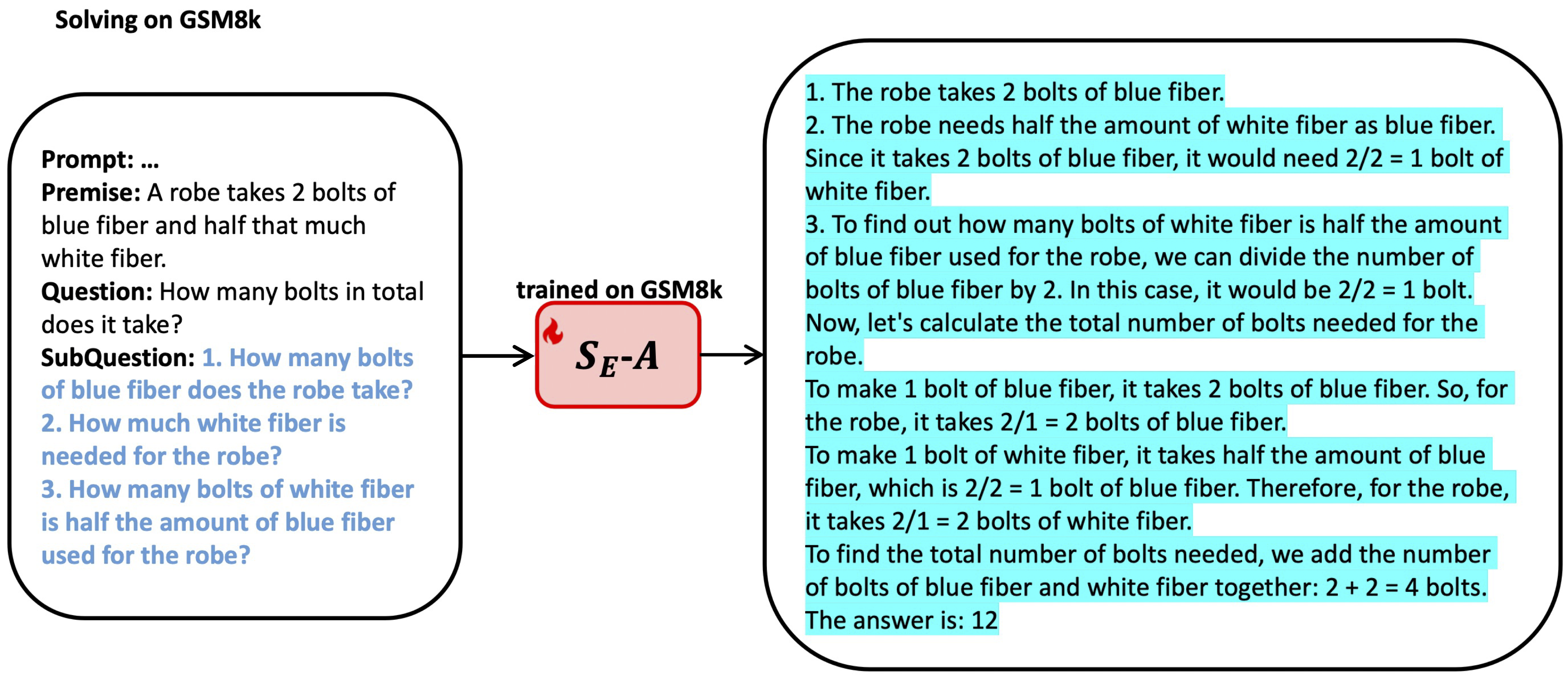}
		\caption[7.5pt]{
            Solver models get lost sometimes. 
        }
		\label{fig:example1}
		\vspace{-0.2in}
	\end{center}
\end{figure*}

\section{Examples Where Solver Models Become Confounded by Subquestions}
\label{sec:example1}
As illustrated in Figure~\ref{fig:example1}, up to the second subquestion, the solver model accurately responds that "The robe requires 2 bolts of blue fiber" and "it would need 1 bolt of white fiber." Nevertheless, the introduction of the third subquestion, closely resembling the second, leads to confusion. Consequently, the model deviates from its initial accuracy, culminating in an incorrect answer following this subquestion.

\section{Extended Related Work}
\paragraph{Planning and Task Decomposition of LLM-powered Agent} 
Recent advances in LLM-powered systems have made it possible to create an end-to-end pipeline, opening up new possibilities for developing autonomous agents that can complete complex tasks using enhanced planning and memory capabilities. 
Promising works, such as ReAct \citep{yao2022react}, HuggingGPT \citep{shen2023hugginggpt}, AutoGPT \citep{Auto-GPT}, LangChain \citep{langchain-ai}, GPT-Engineer \citep{gpt-engineer}, BabyAGI \citep{babyagi} and CoMM \citep{chen-etal-2024-comm}, have demonstrated significant potential in this field.
These agents rely on the LLM to decompose larger tasks into more manageable components.
Among them, some approaches (\textit{e.g.}, HuggingGPT) use a \textit{static} planning strategy by first generating the complete plan via LLM and subsequently tackling each subtask. Other approaches (\textit{e.g.}, AutoGPT) adopt a \textit{dynamic} and \textit{interactive} planning strategy, where the generation of each action is conditioned on the outcome of the previous planning steps.

\end{document}